\documentclass[runningheads,a4paper]{llncs}

\usepackage{graphicx}

\usepackage{placeins}

\usepackage[latin1]{inputenc}

\usepackage{amsfonts}
\usepackage{amssymb}
\usepackage{scalerel}

\usepackage{url}    
\newcommand{\keywords}[1]{\par\addvspace\baselineskip
\noindent\keywordname\enspace\ignorespaces#1}

\newcommand\scale[2]{\vstretch{#1}{\hstretch{#1}{#2}}}
\newcommand{\LIPplus}{\mathbin{\ooalign{$\bigtriangleup$\crcr\hidewidth
  \raise.14em\hbox{$\scale{0.7}{\scriptscriptstyle+}$}\hidewidth}}}
\newcommand{\LIPminus}{\mathbin{\ooalign{$\bigtriangleup$\crcr\hidewidth
  \raise.14em\hbox{$\scale{0.7}{\scriptscriptstyle-}$}\hidewidth}}}
\newcommand{\LIPtimes}{\mathbin{  \ooalign{$\bigtriangleup$\crcr\hidewidth
  \raise.14em\hbox{$\scale{0.7}{\scriptscriptstyle\times}$}\hidewidth}}}
\newcommand{\LIPCplus}{\LIPplus_{c}}

\newcommand{\LIPCtimes}{\LIPtimes_{c}}

\newcommand{\Real}{\mathbb R}
\newcommand{\Tcurv}{\mathcal{T}}

\newcommand{\fx}{\mathbf{f}(x)}
\newcommand{\gx}{\mathbf{g}(x)}
\newcommand{\la}{\lambda}
\newcommand{\F}{\mathbf{F}}
\newcommand{\G}{\mathbf{G}}
\newcommand{\K}{\mathbf{\acute{K}}}
\newcommand{\U}{\mathbf{\acute{U}}}
\newcommand{\T}{\mathbf{T}}
\newcommand{\C}{\mathbf{C}}

\newcommand{\I}{\mathcal{I}^3}

\begin{document}

\mainmatter  

\title{Spatio-colour Aspl\"und's metric and Logarithmic Image Processing for Colour Images (LIPC)}

\titlerunning{Spatio-colour Aspl\"und's metric with LIPC model}

%
%
\author{Guillaume Noyel\inst{1}%
\and Michel Jourlin\inst{1,2}}
\authorrunning{Guillaume Noyel et al.}

\institute{International Prevention Research Institute,	95 cours Lafayette,	69006 Lyon, France
\and Lab. H. Curien, UMR CNRS 5516,	18 rue Pr. B. Lauras,	42000 St-Etienne, France\\
\url{www.i-pri.org}}

%
%

\toctitle{Spatio-colour Aspl\"und's metric and Logarithmic Image Processing for Colour Images (LIPC)}
\tocauthor{Guillaume Noyel, Michel Jourlin}
\maketitle

\begin{abstract}
Aspl\"und's metric, which is useful for pattern matching, consists in a double-sided probing, i.e. the over-graph and the sub-graph of a function are probed jointly. This paper extends the Aspl\"und's metric we previously defined for colour and multivariate images using a marginal approach (i.e. component by component) to the first spatio-colour Aspl\"und's metric based on the vectorial colour LIP model (LIPC). LIPC is a non-linear model with operations between colour images which are consistent with the human visual system. The defined colour metric is insensitive to lighting variations and a variant which is robust to noise is used for colour pattern matching.
\end{abstract}

\section{Introduction}
\label{sec:intro}

The Aspl\"und's metric initially defined for binary shapes \cite{Asplund1960,Grunbaum1963} has been extended to grey-scale
images by Jourlin et al. \cite{Jourlin2012,Jourlin2014} and to colour and multivariate images in the LIP framework by Noyel et al. \cite{Noyel2015}. It consists in probing a function by two homothetic template functions, i.e. the probes which are computed by the LIP multiplication.

The Logarithmic Image Processing (LIP) model initially defined for grey level images by Jourlin et al. \cite{Jourlin1988,Jourlin2001} is perfectly suited for images acquired by transmitted light (i.e. when the observed object is located between the source and the sensor) and by reflected light because of its consistency with the Human Vision \cite{Brailean1991}. The necessity to analyse together the channels of the colour images (i.e. by a vectorial analysis) has led to the introduction of the Logarithmic Image Processing for Colour images (LIPC) by Jourlin et al. \cite{Jourlin2011}.

The LIP Aspl\"und's metric was defined in \cite{Noyel2015} in a marginal way (i.e. channel by channel). In this paper, our contribution is to extend this metric by using the spatio-colour properties \cite{Noyel2007,Noyel2014} of the colour LIPC framework.

After some prerequisites about the colour LIPC model and about the marginal LIP Aspl\"und's metric, we will define a spatio-colour Aspl\"und's metric in the LIPC framework. Then we will perform spatio-colour pattern matching which is robust to noise. Examples will illustrate the definitions.


%
%

\section{Prerequisites}
\label{sec:pre}


\subsection{LIPC model}
\label{ssec:LIPC}

A colour image $\mathbf{f}$, defined on a domain $D \subset \mathbb{R}^{N}$, with values in $\Tcurv^3 = [0,M[^3$, $M \in \Real$, is written:
\begin{equation}\label{eq:f_col}
	\mathbf{f} : \left\{
	\begin{array}{ccc}
		D &\rightarrow& \Tcurv^3 = [0,M[^3\\
		x &\rightarrow& \mathbf{f}(x) = \left( f_R(x) , f_G(x) , f_B(x) \right)\\
	\end{array}
	\right.
\end{equation}
$f_R$ , $f_G$ , $f_B$ are the red, green and blue channels (i.e. components) of $\mathbf{f}$, $\mathbf{f}(x)$ is a vector-pixel and
$x$ is the spatial coordinate of the vector-pixel. The real value $M$ is equal to $2^8=256$ for 8 bits images. Given $P$ the number of pixels, the matrix $\F$ of $E \rightarrow \Tcurv$, $E = 3 \times P$, associated to the image $\mathbf{f}$ is written:
\begin{equation}\label{eq:F}
	\F = \left[
			\begin{array}{cccc}
				f_R(x_1) & f_R(x_2) & ... & f_R(x_P)\\
				f_G(x_1) & f_G(x_2) & ... & f_G(x_P)\\
				f_B(x_1) & f_B(x_2) & ... & f_B(x_P)\\
			\end{array}
	\right]
\end{equation}
To make the comments easier, the word ``image'' designates both the matrix $\F$ and the image $\mathbf{f}$.
The image space for 24-bits images $\F$ is written $\I$.

A colour image is a particular case of a multivariate image defined as $\mathbf{f}_{\lambda} : D \rightarrow \Tcurv^L$, where $L \in \mathbb{N}$ is the number of channels \cite{Noyel2007,Noyel2014}.

As for the grey-level LIP, the colour LIPC framework is based on colour transmittance \cite{Jourlin2011}. It is valid for transmitted and reflected images \cite{Brailean1991}. It models the human perceptual system approach by taking into account: $i)$ the sensitivity of the human eye in the visible domain characterised by colour matching functions of Stiles and Burch (1959) \cite{Stiles1959} and $ii)$ the spectral distribution of light with the D65 illuminant \cite{Schanda2007}.


In the LIPC framework, the transmittance of the sum of two images $\T_{\F \LIPCplus \G}$ is equal to the product of their transmittances $\T_{\F}$ and $\T_{\G}$: $\T_{\F \LIPCplus \G} = \T_{\F} * \T_{\G}$. The symbol of the LIPC addition is $\LIPCplus$ and $*$ represents the element-wise multiplication \cite{Jourlin2011}. The addition of two images $\F, \G \in \I$ is:
\begin{equation}\label{eq:F+G}
	\F \LIPCplus \G = \K^{-1} \U ( \U^{-1} \K \F * \U^{-1} \K \G).
\end{equation}
$\K$ and $\U$ are real matrices of size $3 \times 3$ corresponding to the LIPC mixing model 
\footnote{With colour matching functions of Stiles and Burch (1959) and D65 illuminant \cite{Jourlin2011}, matrices $\K$ and $\U$ equal to:\\
\begin{equation}\label{eq:pre:U_K}
\begin{array}{@{}l@{ }l@{}}
	\U = \left[ \begin{array}{@{}r@{}c@{}l@{\hspace{1mm}}r@{}c@{}l@{\hspace{1mm}}r@{}c@{}l@{}}
		25&.&0440  & 53&.&1416  & 176&.&8144\\
    21&.&3002  & 185&.&9744 & 47&.&7254\\
    229&.&2474 & 19&.&9944  & 5&.&7583
	\end{array}
	\right]
&
	\K = \left[ \begin{array}{@{}r@{}c@{}l@{\hspace{1mm}}r@{}c@{}l@{\hspace{1mm}}r@{}c@{}l@{}}
		0&.&6991 & 0&.&2109 & 0&.&0899\\
		0&.&1947 & 0&.&8002 & 0&.&0049\\
		0&.&0681 & 0&.&0002 & 0&.&9315  
	\end{array}
	\right]
\end{array}
\end{equation}
}.
From the LIPC addition, a multiplication by a scalar $\alpha \in \Real$ has been defined:
\begin{equation}\label{eq:a*F}
	\alpha \LIPCtimes \F = \K^{-1} \U ( \U^{-1} \K \F )^{\alpha}.
\end{equation}
The space $(\I,\LIPCplus,\LIPCtimes)$ is the positive cone of a vector space with robust mathematical properties.

Physical interpretation \cite{Jourlin2011}: the LIPC addition corresponds to the superposition of two semi-transparent layers. A LIPC multiplication by a scalar $\alpha \in ]0,1[$ brightens the result by suppressing layers, while a scalar $\alpha \in ]1,+\infty[$ darkens the result by superimposing $\alpha$ times the image on itself.


\subsection{Marginal Aspl\"und's metric for colour and multivariate images}
\label{ssec:As_LIP}

In \cite{Noyel2015}, an Aspl\"und's metric between colour images was defined with the LIP model by using a marginal approach (i.e. channel by channel) \cite{Noyel2007,Noyel2014} .
\begin{definition}
The Aspl\"und's metric (with LIP multiplication) between two colour images $\mathbf{f}$ and $\mathbf{g}$ on a region $Z \subset D$ is 
\begin{equation}\label{eq:d_As_C}
d_{As,Z}^{\LIPtimes} (\mathbf{f},\mathbf{g}) = \ln( \lambda / \mu )
\end{equation}
with $\lambda = \inf \left\{ k, \forall x \in Z , k \LIPtimes g_R(x) \geq f_R(x), k \LIPtimes g_G(x) \geq f_G(x) , k \LIPtimes g_B(x) \geq f_B(x) \right\}$ \\
and $\mu      = \sup \left\{ k, \forall x \in Z , k \LIPtimes g_R(x) \leq f_R(x), k \LIPtimes g_G(x) \leq f_G(x) , k \LIPtimes g_B(x) \leq f_B(x) \right\}$.
\end{definition}
In particular, by the property of the distance $d_{As,Z}^{\LIPtimes} (\mathbf{f},\mathbf{g}) = d_{As,Z}^{\LIPtimes} (\mathbf{g},\mathbf{f})$.


%
%
\section{Aspl\"und's metric defined in the Logarithmic Image Processing Colour (LIPC) framework}
\label{sec:As_LIPC}

Given two colours $C_1=(r_1,g_1,b_1)$, $C_2=(r_2,g_2,b_2) \in \mathcal{T}^3$, as we are only looking for lower and upper bounds, a marginal order \cite{Barnett1976} is used:	$C_1 \geq C_2 \Leftrightarrow \{r_1\geq r_2$ and $g_1\geq g_2$ and $b_1\geq b_2 \}$.

\begin{definition}
Given two colours $C_1$, $C_2 \in \mathcal{T}^3$, their Aspl\"und's distance (with LIPC multiplication) is equal to:
\begin{equation}\label{eq:d_As_C1C2}
 d_{As}^{\LIPCtimes} ( C_1 , C_2 ) =  \ln( \mu / \la )
\end{equation}
$\la = \inf_k \left\{ k \LIPCtimes C_2 \geq C_1 \right\}$ and $\mu = \sup_k \left\{ k \LIPCtimes C_2 \leq C_1 \right\}$.
\end{definition}

Strictly speaking, $d_{As}^{\LIPCtimes}$ is a metric if the colours $C_n$ are replaced by their equivalence classes $\tilde{C}_n = \left\{ C \in \mathcal{T}^3 / \exists \alpha \in \Real^{+},\right.$ $\left. \alpha \LIPCtimes C = C_n \right\}$.

Comment: in eq. \ref{eq:d_As_C1C2} contrary to the Aspl\"und's distance (with LIP multiplication) defined in \cite{Noyel2015} (eq. \ref{eq:d_As_C}), we have $\la \leq \mu$ because, by definition of the colour LIPC model the scales are inverted as compared to the grey LIP model \cite{Jourlin2011}.


Colour metrics (with LIPC multiplication) between two colour images $\mathbf{f}$ and $\mathbf{g}$ may be defined as the sum ($d_1$ metric) or the supremum ($d_{\infty}$) of $d_{As}^{\LIPCtimes} ( C_1 , C_2 )$ on the region of interest $Z \subset D$ of cardinal $\#Z$
\begin{equation}
\begin{array}{ccc}
	d_{1,Z}^{\LIPCtimes}(\mathbf{f},\mathbf{g}) &=& \frac{1}{\#Z} \sum_{x \in Z} d_{As}^{\LIPCtimes} (\mathbf{f}(x),\mathbf{g}(x)) \\
	d_{\infty,Z}^{\LIPCtimes}(\mathbf{f},\mathbf{g}) &=& \sup_{x \in Z} d_{As}^{\LIPCtimes} (\mathbf{f}(x),\mathbf{g}(x))
\end{array}
\end{equation}
The Aspl\"und's metric can be extended to colour functions.


\begin{definition}
The colour Aspl\"und's metric (with LIPC multiplication) between two colour images $\mathbf{f}$ and $\mathbf{g}$ on a region $Z \subset D$ is
\begin{equation}\label{eq:d_As}
d_{As,Z}^{\LIPCtimes} (\mathbf{f},\mathbf{g}) = \ln( \mu / \la )
\end{equation}
$\la = \inf_k \left\{ \forall x \in Z , k \LIPCtimes \gx \geq \fx \right\}$ and $\mu  = \sup_k \left\{ \forall x \in Z , k \LIPCtimes \gx \leq \fx \right\}$.
\end{definition}

\begin{figure}[!htb]
\begin{tabular}{@{}c@{}c@{}c@{}}
\includegraphics[width=0.3\columnwidth]{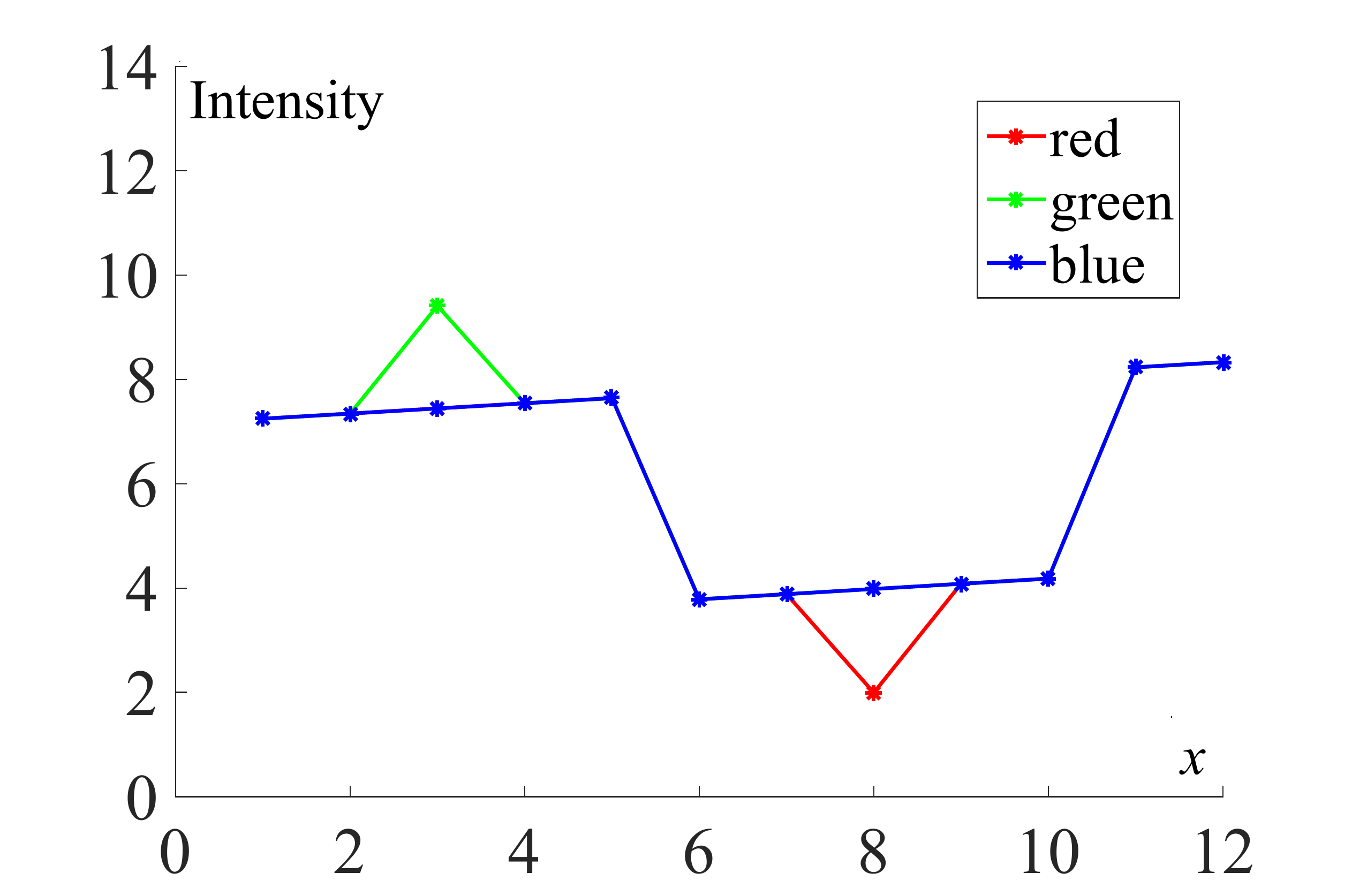}&
\includegraphics[width=0.3\columnwidth]{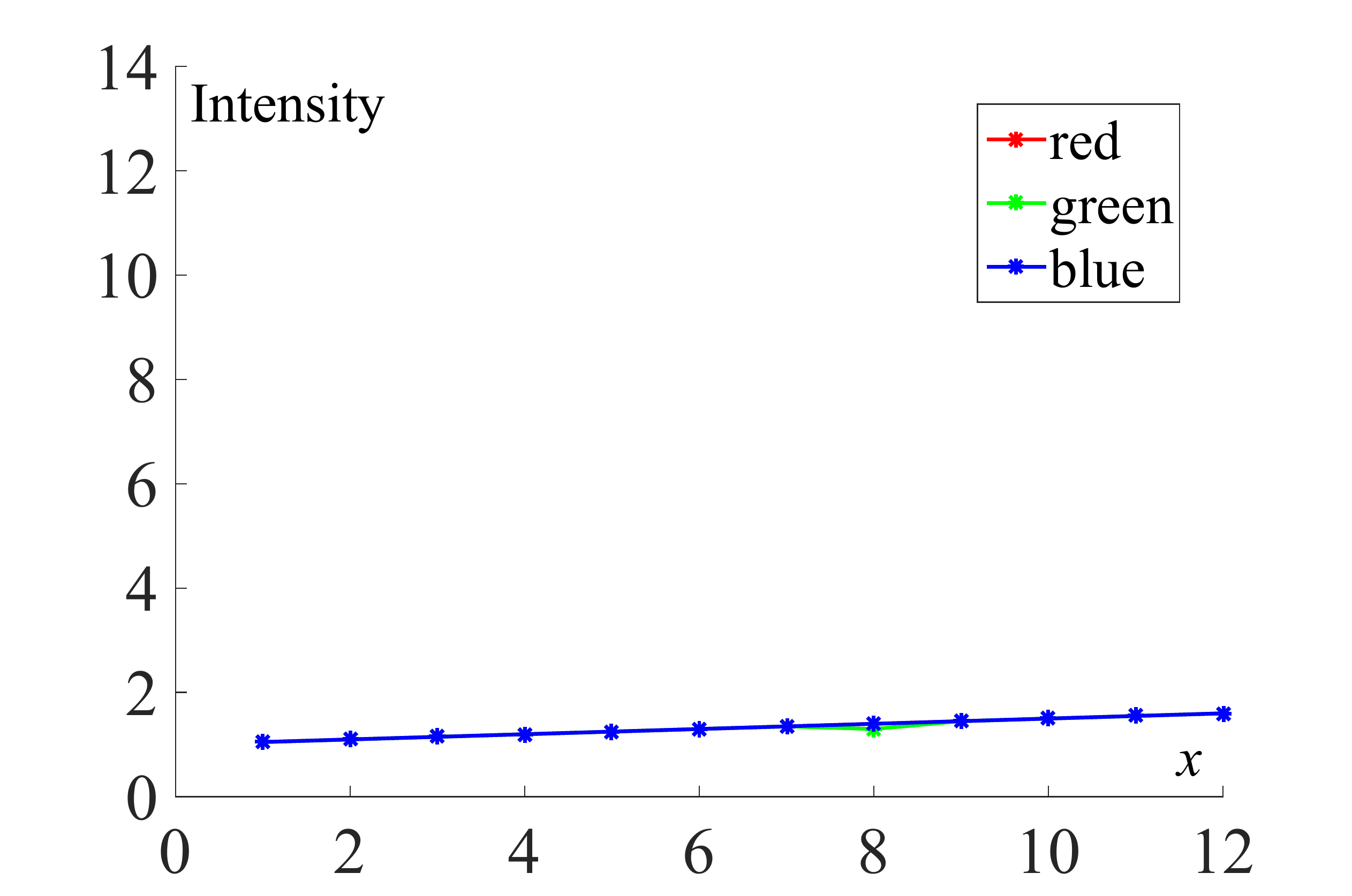}&
\includegraphics[width=0.3\columnwidth]{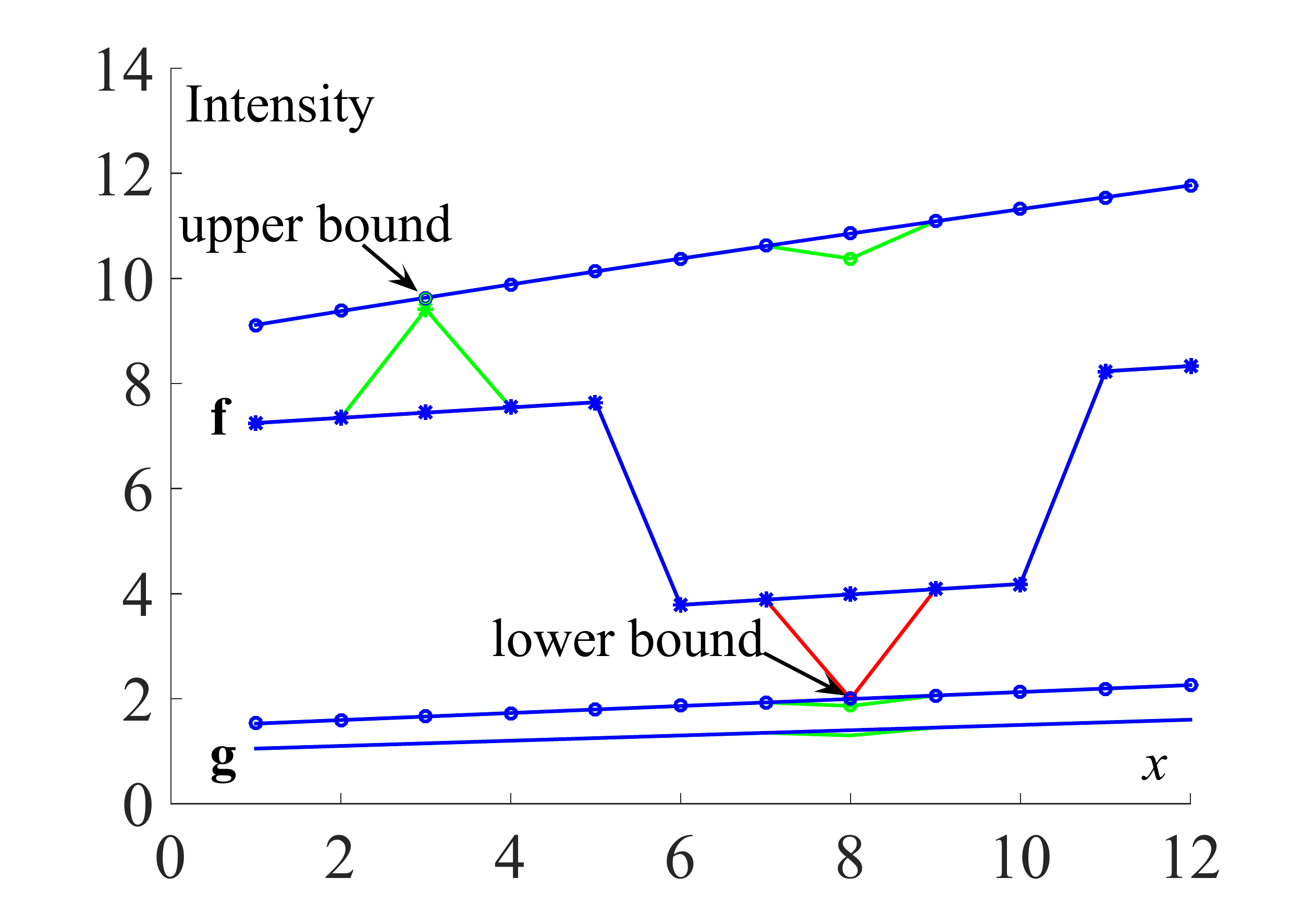}\\
(a) Colour function $\mathbf{f}$&
(b) Colour probe $\mathbf{g}$&
(c) Lower ($\mu$) and upper ($\lambda$)	bounds
\end{tabular}
\caption{Computation of the Aspl\"und's distance between two colour functions 
$d_{As,D}^{\protect \LIPCtimes}(\mathbf{f},\mathbf{g}) = 0.43$. 
Each colour channel is represented by a line of the same colour.}
\label{fig:das_v_sig}
\end{figure}

In fig. \ref{fig:das_v_sig}, the Aspl\"und's metric has been computed between the colour probe $\mathbf{g}$ and the colour function $\mathbf{f}$ on their definition domain $D$. 

Comment: the lower (resp. upper) bound $\mu \LIPCtimes \mathbf{g}$
(resp. $\la \LIPCtimes \mathbf{g}$) may not be equal to any point of the function $\mathbf{f}$ but strictly less (or greater) than the function.
Indeed, one can demonstrate that the following assertion is verified: ``given $\C_0 \in \Tcurv^3, \forall C \in \Tcurv^3, \not \exists \la \in \Real^+ / \la \LIPCtimes C_0 = C$''.


The metric $d_{As,Z}^{\LIPCtimes}$ can be adapted to local processing with a colour template image (i.e. the probe) $\mathbf{t}$ defined on a spatial support $D_t \subset D$. For each point $x \in D$, the distance $d_{As,D_t}^{\LIPCtimes} (\mathbf{f}_{\left|D_t(x)\right.},\mathbf{t})$ is computed on the neighbourhood $D_t(x)$ centred in $x$ where $\mathbf{f}_{\left|D_t(x)\right.}$ is the restriction of $\mathbf{f}$ to $D_t(x)$. 

\begin{definition}
Given a colour image $\mathbf{f}$ defined on $D$ with values in $\Tcurv^3$, $\left(\Tcurv^{3}\right)^{D}$, a colour probe $\mathbf{t}$ defined on $D_t$  with values in $\Tcurv^{3}$, $\left(\Tcurv^{3}\right)^{D_{t}}$, and $D_t(x)$ the neighbourhood $D_t$ centred in $x \in D$, the map of Aspl\"und's distances (with $\LIPCtimes$) is:
\begin{equation}\label{eq:As_v_C}
	As_{\mathbf{t}}^{\LIPCtimes}\mathbf{f} : \left\{
	\begin{array}{ccc}
		\left(\Tcurv^{3}\right)^{D} \times \left(\Tcurv^{3}\right)^{D_{t}} &\rightarrow& \left(\Real^{+}\right)^{D}\\
		(\mathbf{f},\mathbf{t}) &\rightarrow& As_{\mathbf{t}}^{\LIPCtimes}\mathbf{f}(x) = d_{As,D_t}^{\LIPCtimes} (\mathbf{f}_{\left|D_t(x)\right.},\mathbf{t})\\
	\end{array}
	\right.
\end{equation}
\end{definition}

\begin{figure}[!htb]
\begin{tabular}{@{}c@{}c@{}c@{}}
\includegraphics[width=0.33\columnwidth]{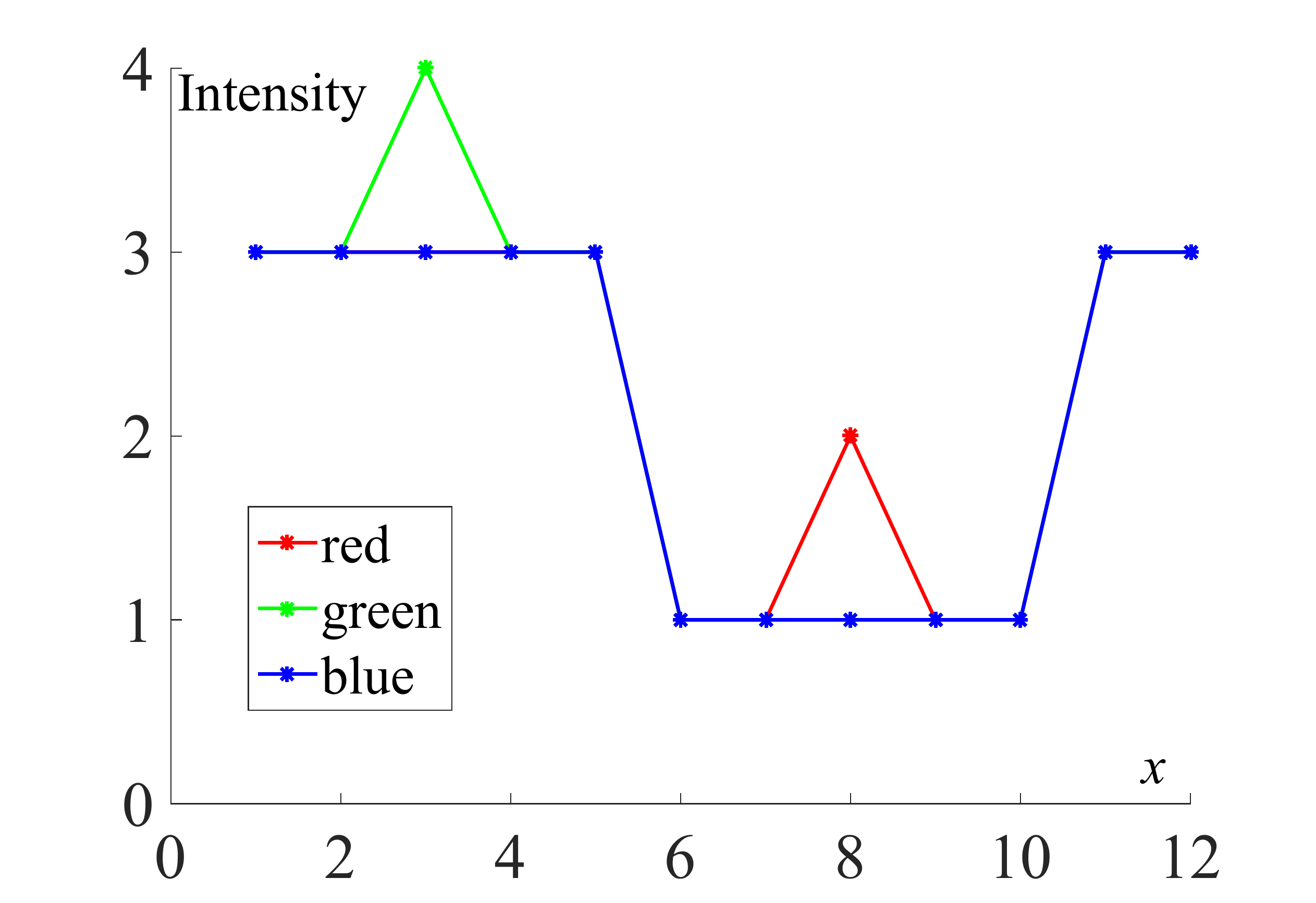}&
\includegraphics[width=0.33\columnwidth]{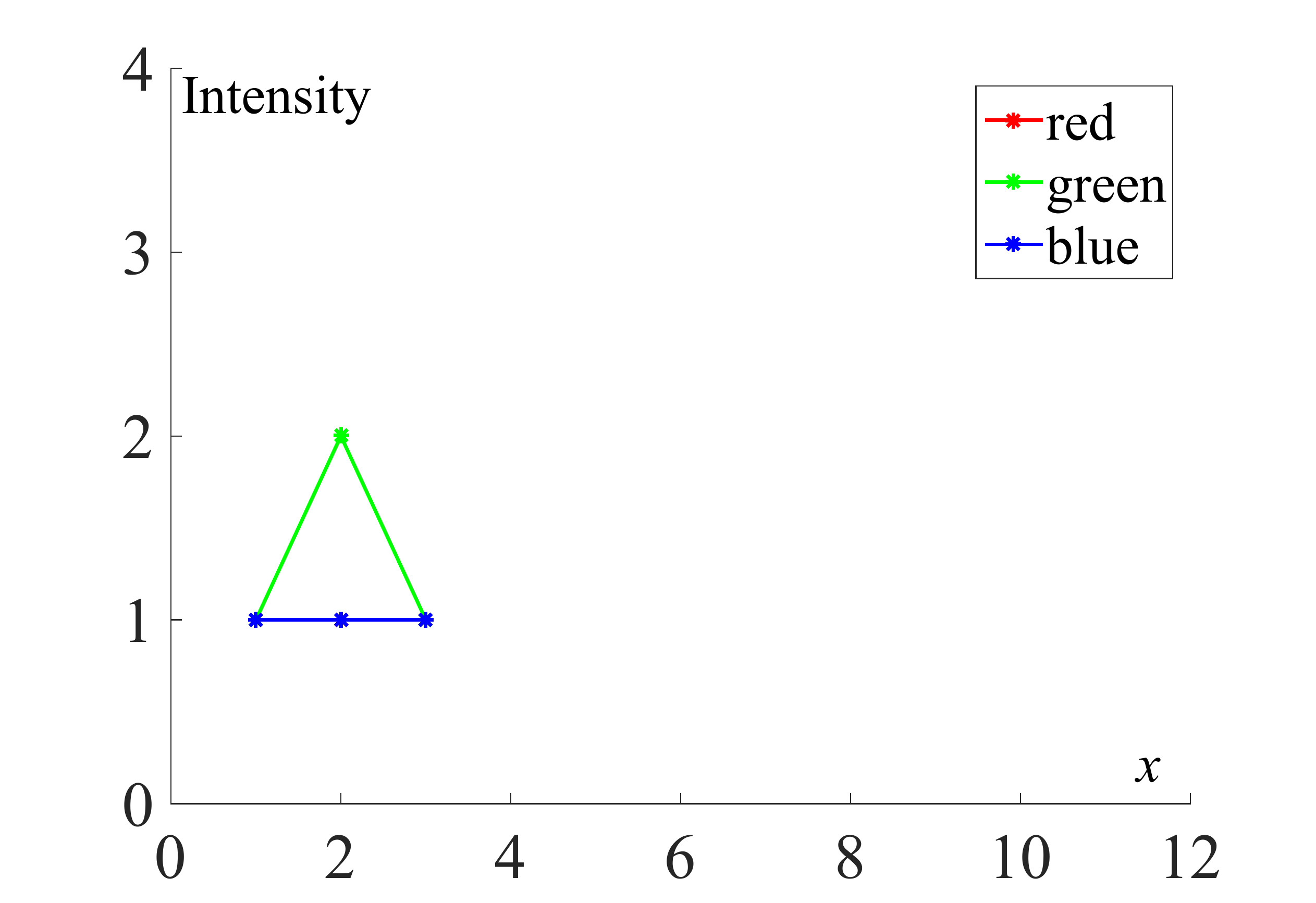}&
\includegraphics[width=0.33\columnwidth]{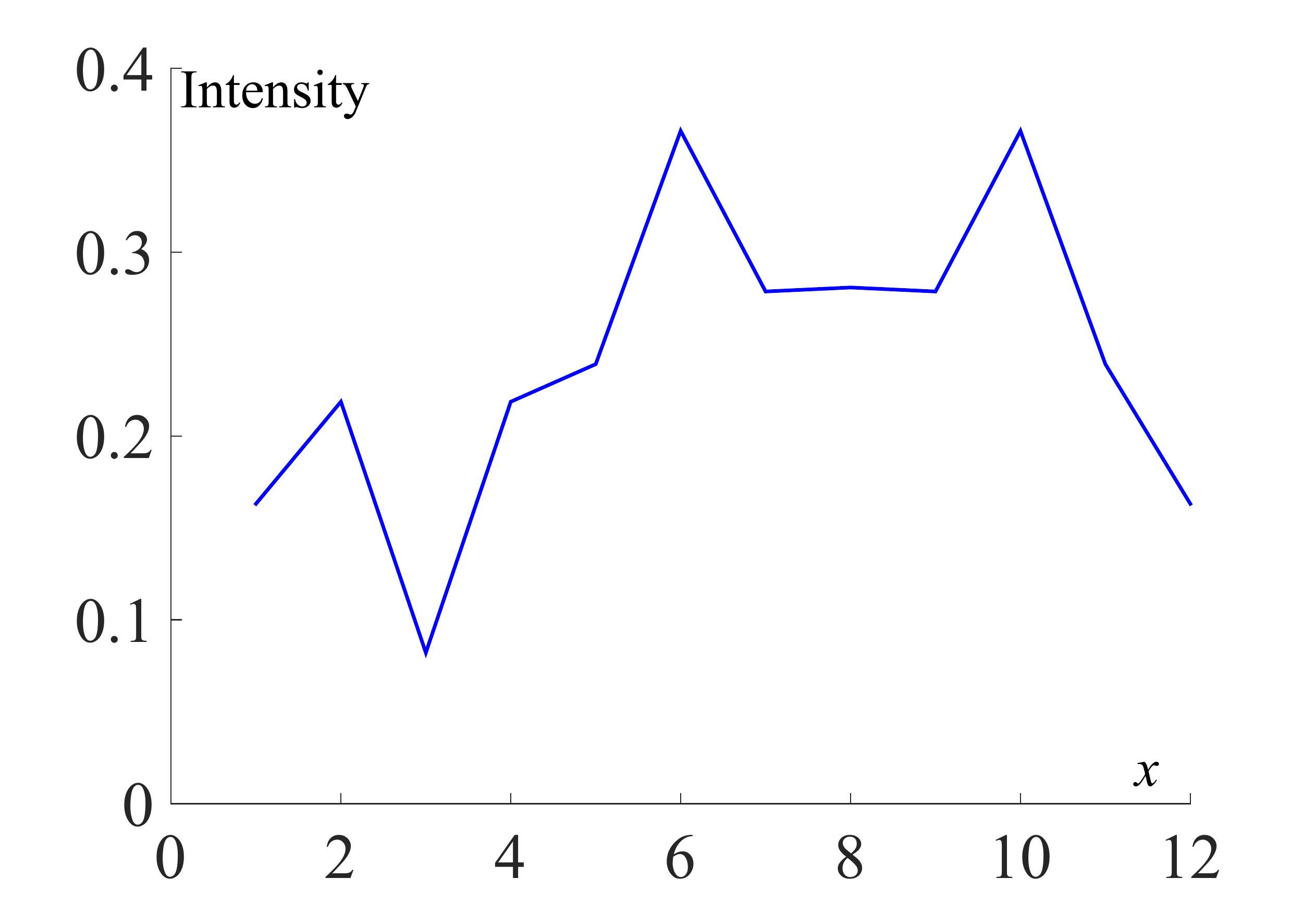}\\
(a) Colour function $\mathbf{f}$&
(b) Colour probe $\mathbf{t}$&
(c) Map of Aspl\"und's\\
	&&distance $As_{\mathbf{t}}^{\LIPCtimes}\mathbf{f}$
\end{tabular}
\caption{(c) Map of the Aspl\"und's distances $As_{\mathbf{t}}^{\protect \LIPCtimes}\mathbf{f}$ between a colour function and a probe. (a) and (b) Each colour channel is represented by a line of the same colour.}
\label{fig:map_as_v_sig}
\end{figure}

In figure \ref{fig:map_as_v_sig}, the map of Aspl\"und's distances is computed between a colour function and a colour probe. The minima of the map corresponds to the location of a pattern which is similar to the probe.

Aspl\"und's distance is sensitive to noise because the probe lays on regional extrema that may be caused by noise (Figure \ref{fig:das_v_sig}). In \cite{Jourlin2014,Noyel2015}, definitions of Aspl\"und's distance with a tolerance on the extrema have been introduced. In this paper, we extend this definition for colour images with LIPC model. 

To reduce the sensitivity of Aspl\"und's distance to the noise, the ``Measure metric'' or ``M-metric'' has been defined in the context of ``Measure Theory''. The image being digitized, the number of pixels of $D$ is finite and the ``measure'' of a subset of $D$ is linked to the cardinal of this subset, e.g. the percentage $P$ of its elements with respect to $D$. We are looking for a subset $D'$ of $D$, such that $\mathbf{f}_{\left|D'\right.}$ and $\mathbf{g}_{\left|D'\right.}$ are neighbours for Aspl\"und's metric and the complementary set $D\setminus D'$ of $D'$ into $D$ is of small size when compared to $D$. This last condition is written as:
 $P(D \setminus D') = \frac{\#(D \setminus D')}{\#D} \leq p$, 
where $p$ is an acceptable percentage and $\#D$ is the number of elements in $D$.

Given $\epsilon$ a small positive real number, the neighbourhood of function $\mathbf{f}$ is
	\begin{equation}
		N_{P,d_{As},\epsilon,p}(\mathbf{f}) = \left\{ \mathbf{g} \setminus \exists D' \subset D, d_{As,D'}^{\LIPCtimes}(\mathbf{f}_{\left|D'\right.},\mathbf{g}_{\left|D'\right.}) < \epsilon \mbox{ and } \frac{\#(D \setminus D')}{\#D} \leq p\right\}
	\end{equation}

The closest points of the probe to the function are discarded as in \cite{Jourlin2011,Noyel2015}.

\begin{definition}\label{def:dAs_c_m_tol}
Given two constant vector-pixels $\mathbf{c}_{\mu}, \mathbf{c}_{\la} \in \Tcurv^3$, a percentage $p$ of points to be discarded. The colour Aspl\"und's metric (with LIPC multiplication) with tolerance between two colour images $\mathbf{f}$ and $\mathbf{g}$ on a region $Z \subset D$ is
\begin{equation}\label{eq:d_As_c_m_tol}
d_{As,Z,p}^{\LIPCtimes} (\mathbf{f},\mathbf{g}) = \ln( \mu' / \la' )
\end{equation}
$\la' = \inf_k \left\{ \forall x \in Z , k \LIPCtimes \gx \geq \fx - \mathbf{c}_{\la} \right\}$ and $\mu'  = \sup_k \left\{ \forall x \in Z , k \LIPCtimes \gx \leq \right.$ $\left. \fx + \mathbf{c}_{\mu} \right\}$. $\mathbf{c}_{\mu}$ and $\mathbf{c}_{\la}$ are increased such as a percentage $p$ of points is discarded. 
\end{definition}

\begin{figure}[!htb]
\begin{tabular}{@{}c@{}c@{}c@{}}
  \includegraphics[width=0.33\columnwidth]{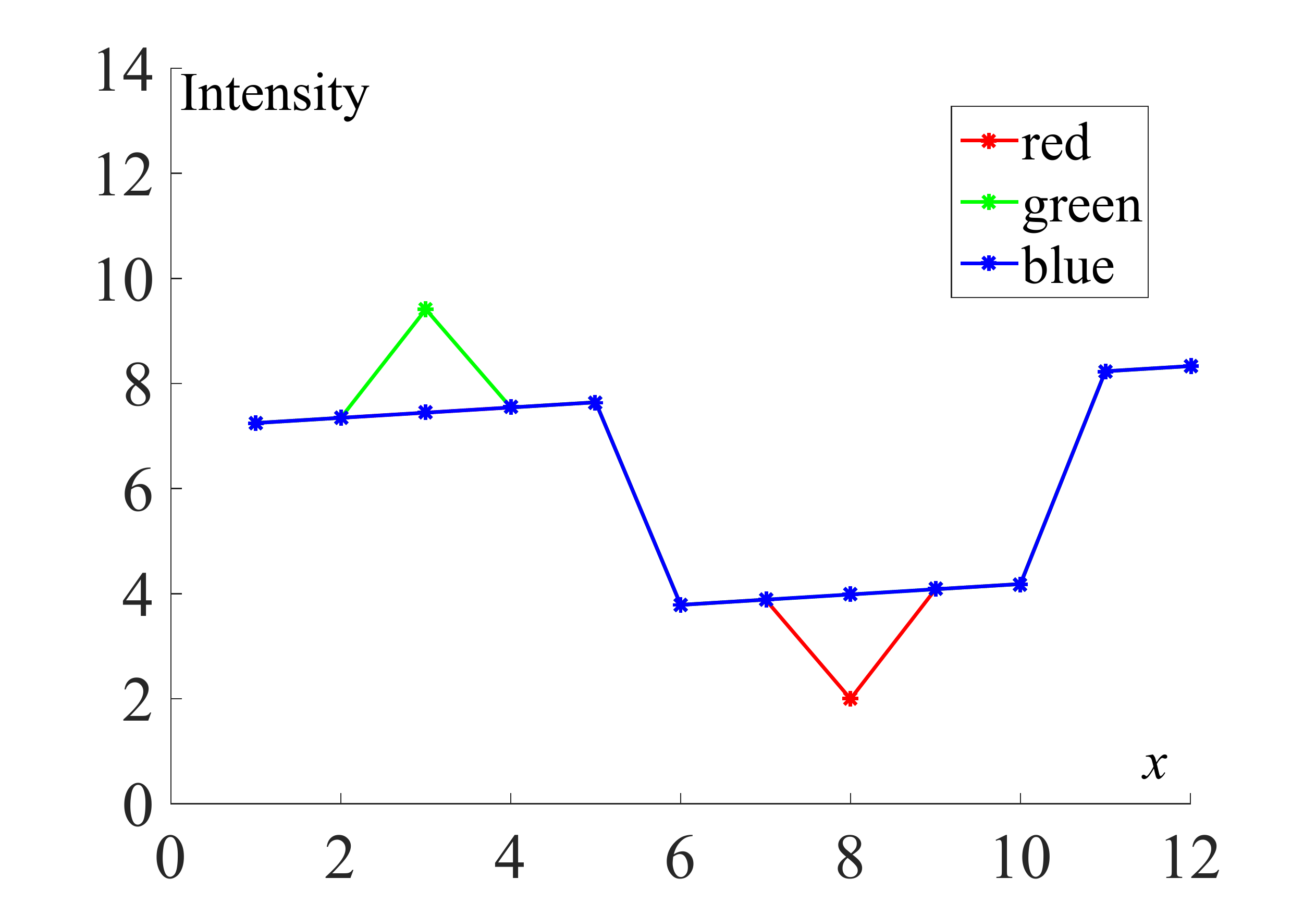}&
  \includegraphics[width=0.33\columnwidth]{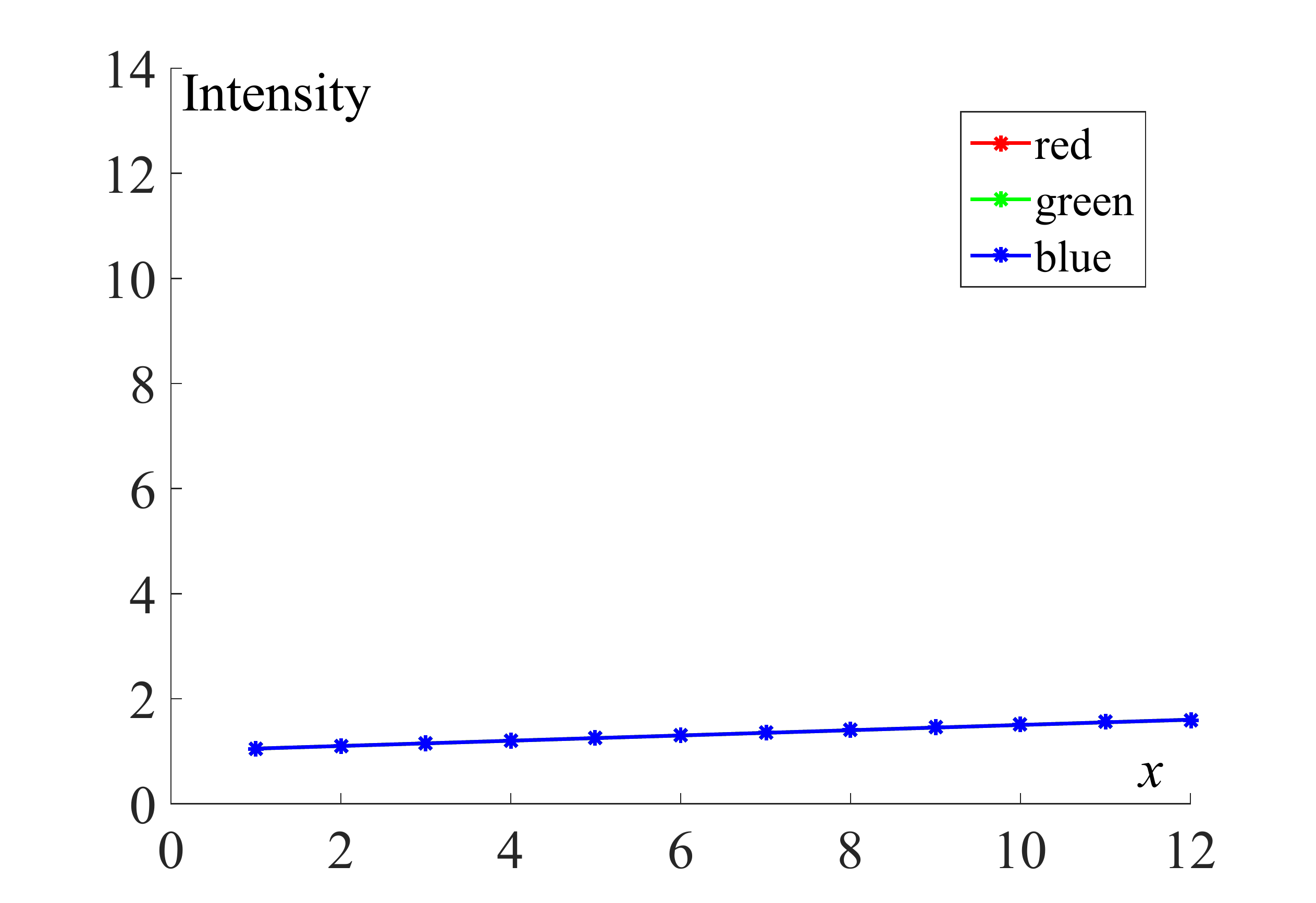}&
	\includegraphics[width=0.33\columnwidth]{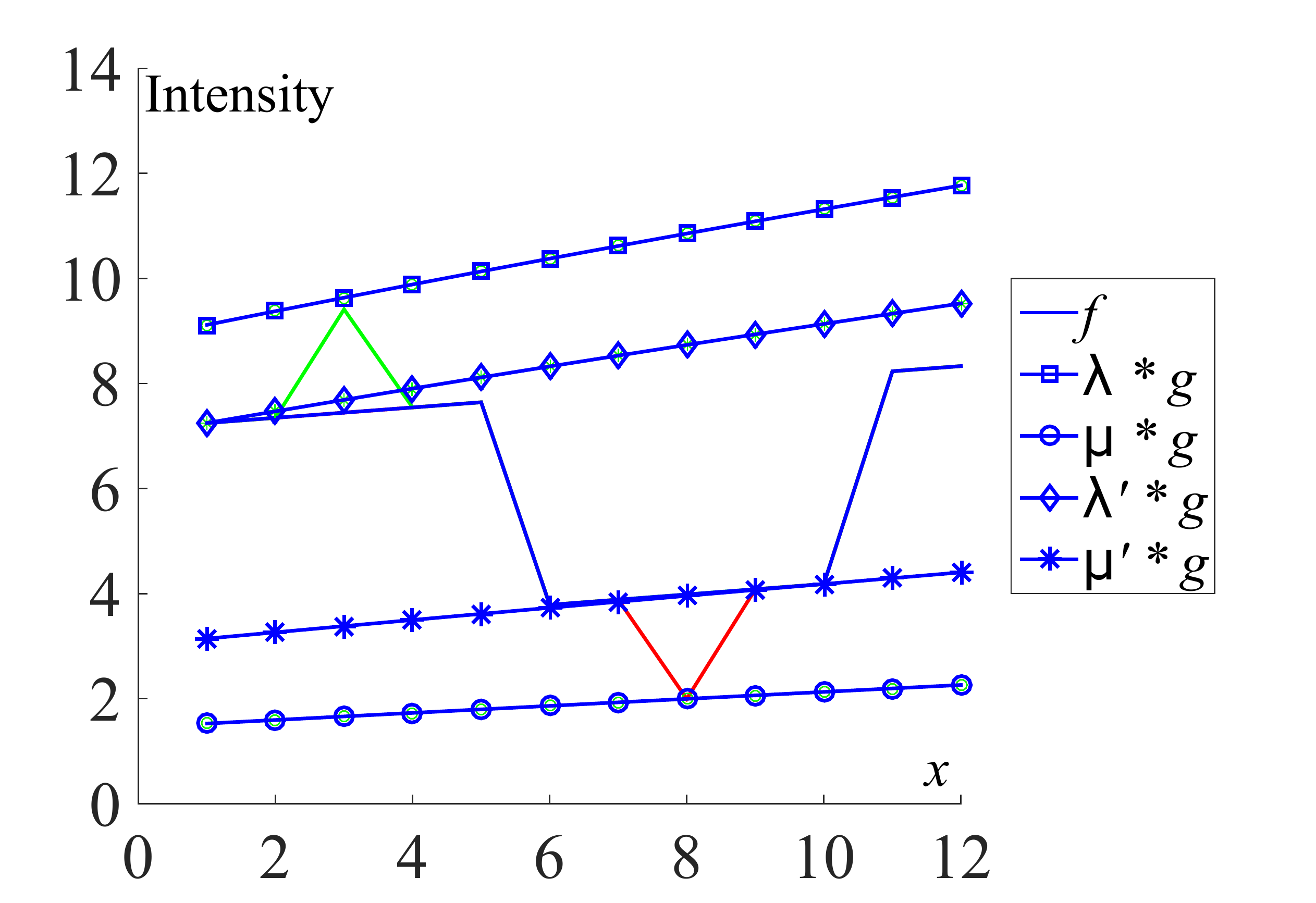}\\
  (a) Colour function $\mathbf{f}$&
	(b) Colour probe $\mathbf{g}$&
	(c) Lower and upper\\
	&&bounds, $p = 20\%$
\end{tabular}
\caption{Colour Aspl\"und's distance with a tolerance of $p = 20 \%$. ($\mu$, $\lambda$) 
are the scalars multiplying the probe without tolerance. ($\mu'$, $\lambda'$) are the scalars multiplying the probe with tolerance. $d_{As,D}^{\protect \LIPCtimes}(\mathbf{f},\mathbf{g}) = 0.43$ and $d_{As,D,p=20\%}^{\protect \LIPCtimes}(\mathbf{f},\mathbf{g}) = 0.21$}
\label{fig:d_as_v_tol}
\end{figure}

In figure \ref{fig:d_as_v_tol}, a tolerance of $p = 20\%$ is used to discard two points. The Aspl\"und's distance decreases from $0.43$ to $0.21$.

A map of Aspl\"und's distances (with $\LIPCtimes$) can now be defined. 
\begin{definition}\label{def:As_c_m_tol}
Given a colour image $\mathbf{f}$ of $\left(\Tcurv^{3}\right)^{D}$, a colour probe $\mathbf{t}$ of $\left(\Tcurv^{3}\right)^{D_{t}}$ and a tolerance $p\in[0,1]$, the map of Aspl\"und's distances with a tolerance is:
\begin{equation}\label{eq:As_C_tol}
	As_{\mathbf{t},p}^{\LIPCtimes}\mathbf{f} : \left\{
	\begin{array}{ccc}
		\left(\Tcurv^{3}\right)^{D} \times \left(\Tcurv^{3}\right)^{D_{t}} &\rightarrow& \left(\Real^{+}\right)^{D}\\
		(\mathbf{f},\mathbf{t}) &\rightarrow& As_{\mathbf{t},p}^{\LIPCtimes}\mathbf{f}(x) = d_{As,D_t,p}^{\LIPCtimes} (\mathbf{f}_{\left|D_t(x)\right.},\mathbf{t})
	\end{array}
	\right.
\end{equation}
$D_t(x)$ is the neighbourhood $D_t$ centred in $x \in D$.
\end{definition}


%
%

\section{Examples}
\label{sec:ex}

\begin{figure}[!htb]
\begin{tabular}{ccc}
\includegraphics[width=0.31\columnwidth]{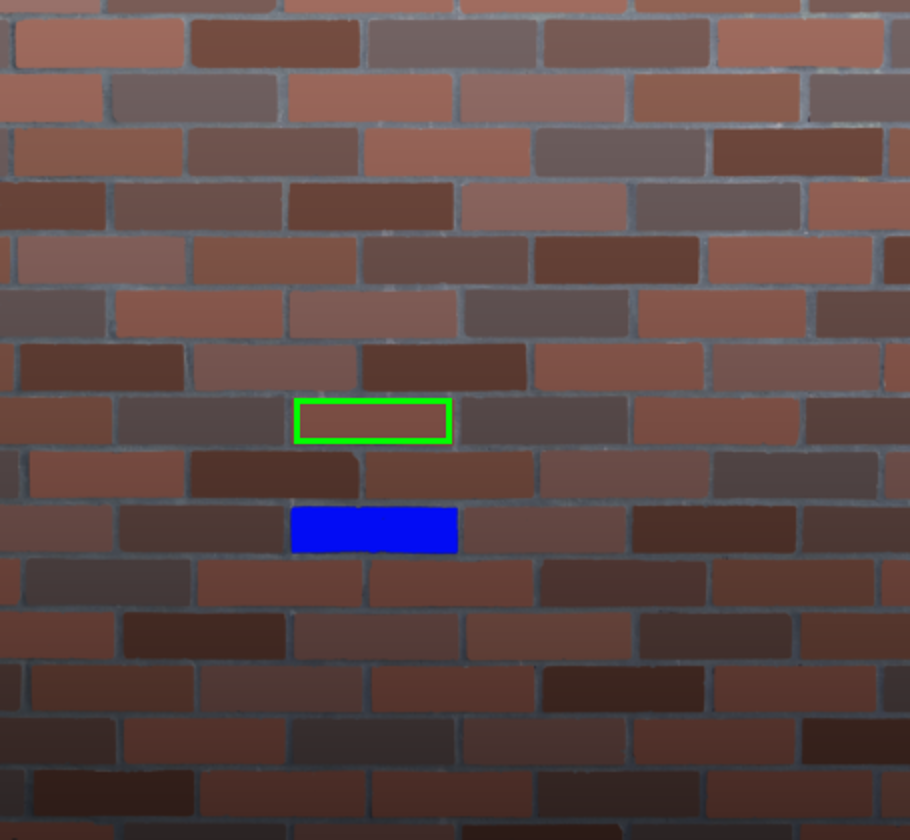}&
\includegraphics[width=0.31\columnwidth]{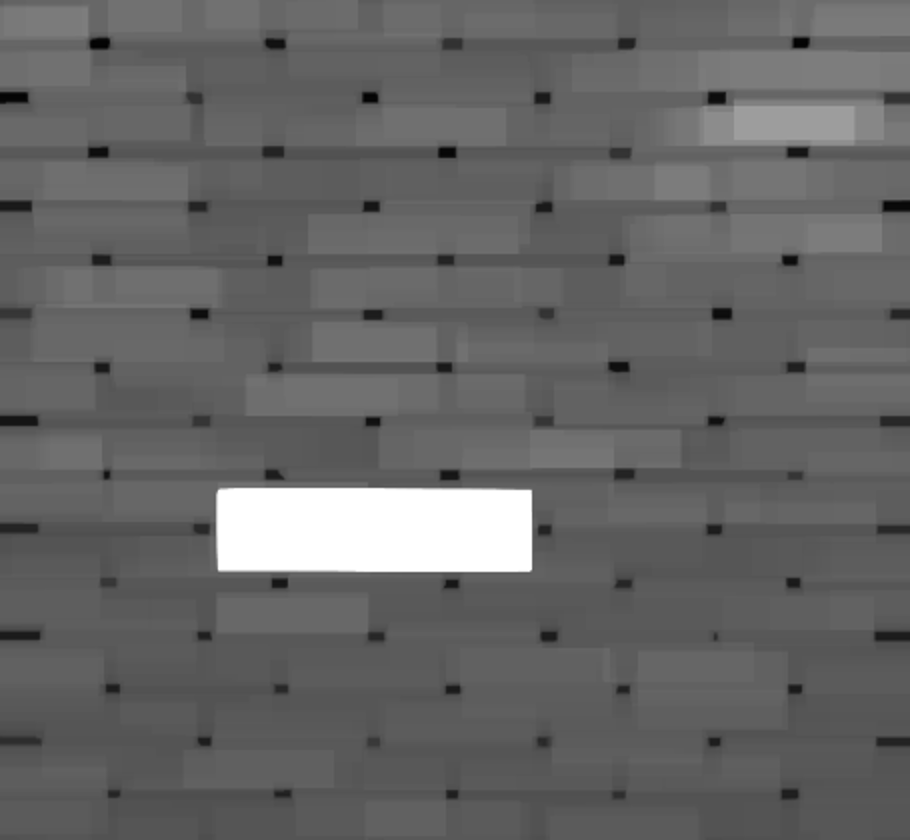}&
\includegraphics[width=0.31\columnwidth]{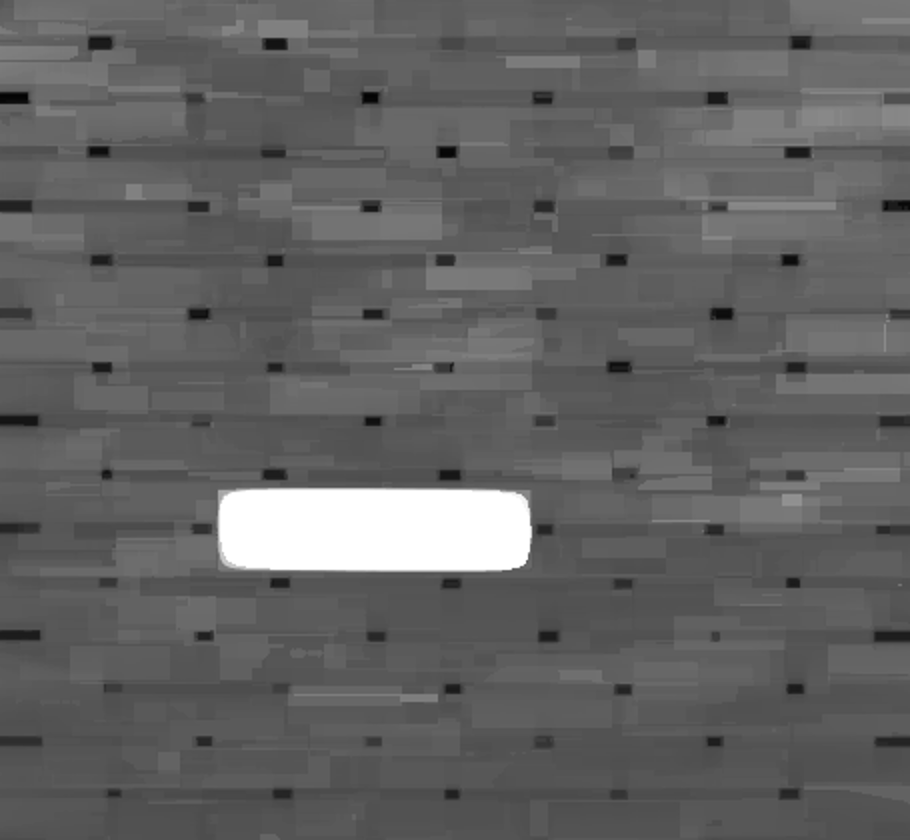}\\
(a) Image $\mathbf{f}$ and probe $\mathbf{t}$&
(b) Map $As_{\mathbf{t}}^{\protect \LIPCtimes}\mathbf{f}$&
(c) Map $As_{\mathbf{t},p=98\%}^{\protect\LIPCtimes}\tilde{\mathbf{f}}$\\

\includegraphics[width=0.31\columnwidth]{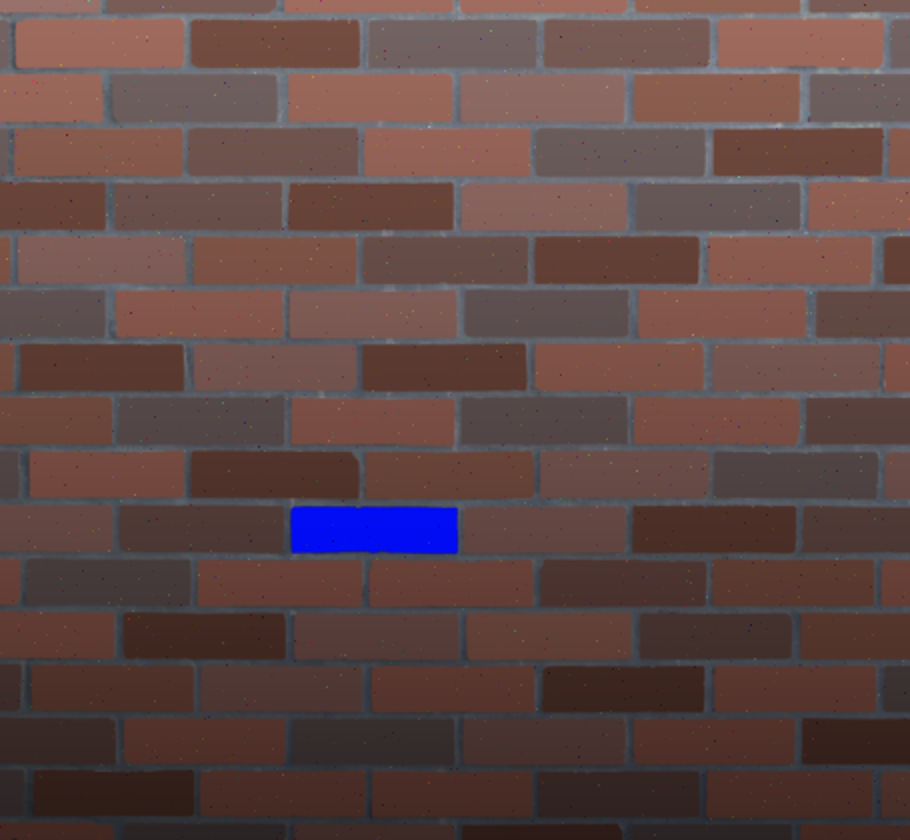}&
\includegraphics[width=0.31\columnwidth]{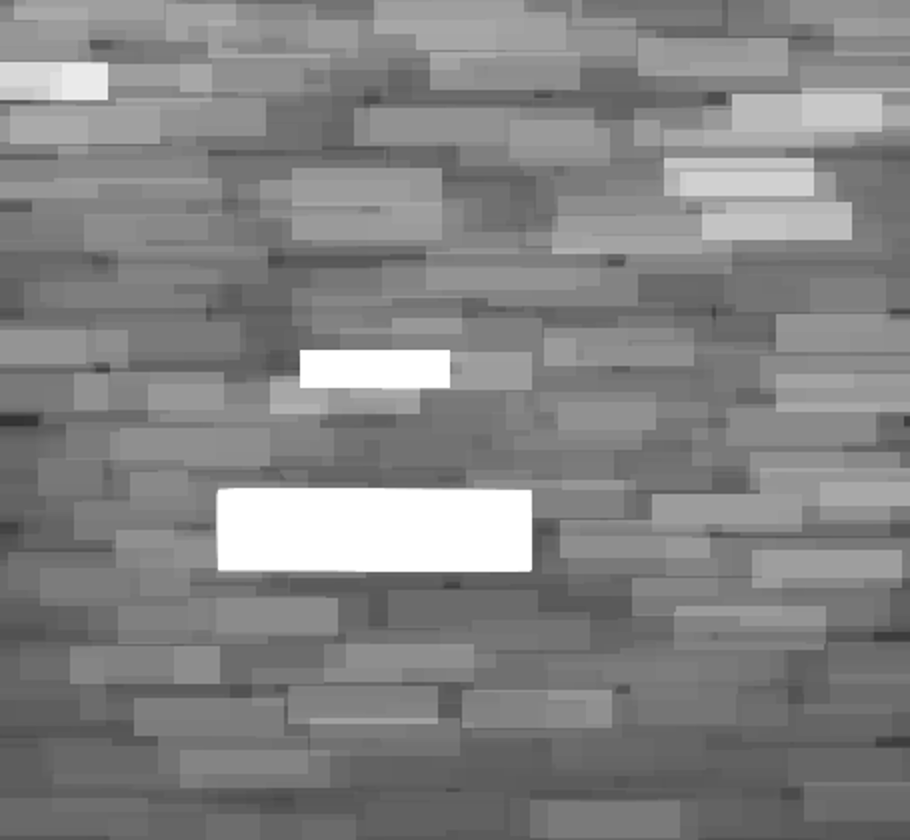}&
\includegraphics[width=0.31\columnwidth]{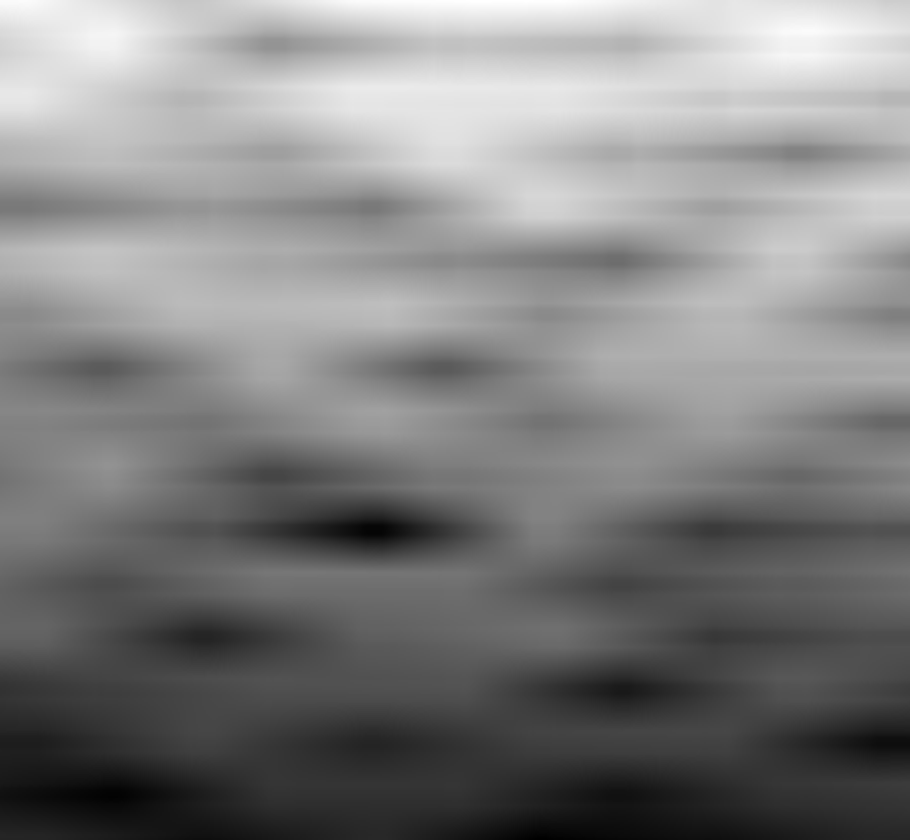}\\
(d) Noisy image $\tilde{\mathbf{f}}$&
(e) Map $As_{\mathbf{t}}^{\protect \LIPCtimes}\tilde{\mathbf{f}}$&
(f) Correlation map
\end{tabular}
\caption{Maps of Aspl\"und's distances without tolerance $As_{\mathbf{t}}^{\protect \LIPCtimes}\tilde{\mathbf{f}}$ 
and with $As_{\mathbf{t},p}^{\protect \LIPCtimes}\tilde{\mathbf{f}}$. $\tilde{\mathbf{f}}$~image with
a white noise ($\sigma^2=2.6$, spatial density $1\%$). (f) Correlation map.}
\label{fig:brick}
\end{figure}
In figure \ref{fig:brick}, we look for the bricks of a wall, similar to a colour probe. A blue brick has been added to the wall. In the image without noise $\mathbf{f}$, the regional minima of the map $As_{\mathbf{t}}^{\LIPCtimes}\mathbf{f}$ (dark points in fig. \ref{fig:brick}b) correspond to the centre of the bricks similar to the probe (according to the Aspl\"und's distance). The white rectangle corresponds to the maxima of the distances between the blue brick and the probe. Therefore, the distance is sensitive to colour (i.e. the hue). In the image with noise $\tilde{\mathbf{f}}$, the map without tolerance $As_{\mathbf{t}}^{\LIPCtimes}\tilde{\mathbf{f}}$ is more sensitive to noise (fig. \ref{fig:brick}e) than the map with tolerance $As_{\mathbf{t},p}^{\LIPCtimes}\tilde{\mathbf{f}}$ (fig \ref{fig:brick}c). Indeed, the minima are preserved into the map with tolerance (fig. \ref{fig:brick}c) compared to the map without (fig. \ref{fig:brick}e). The minima can be extracted using mathematical morphology \cite{Matheron1967,Serra1982}. Importantly, all the maps of Aspl\"und's distances are insensitive to the vertical lighting drift. Moreover, a correlation map is useless to find the location of the bricks (fig. \ref{fig:brick}f).

\begin{figure} 
\begin{tabular}{c}
\begin{tabular}{ccc}
  \includegraphics[width=0.29\columnwidth]{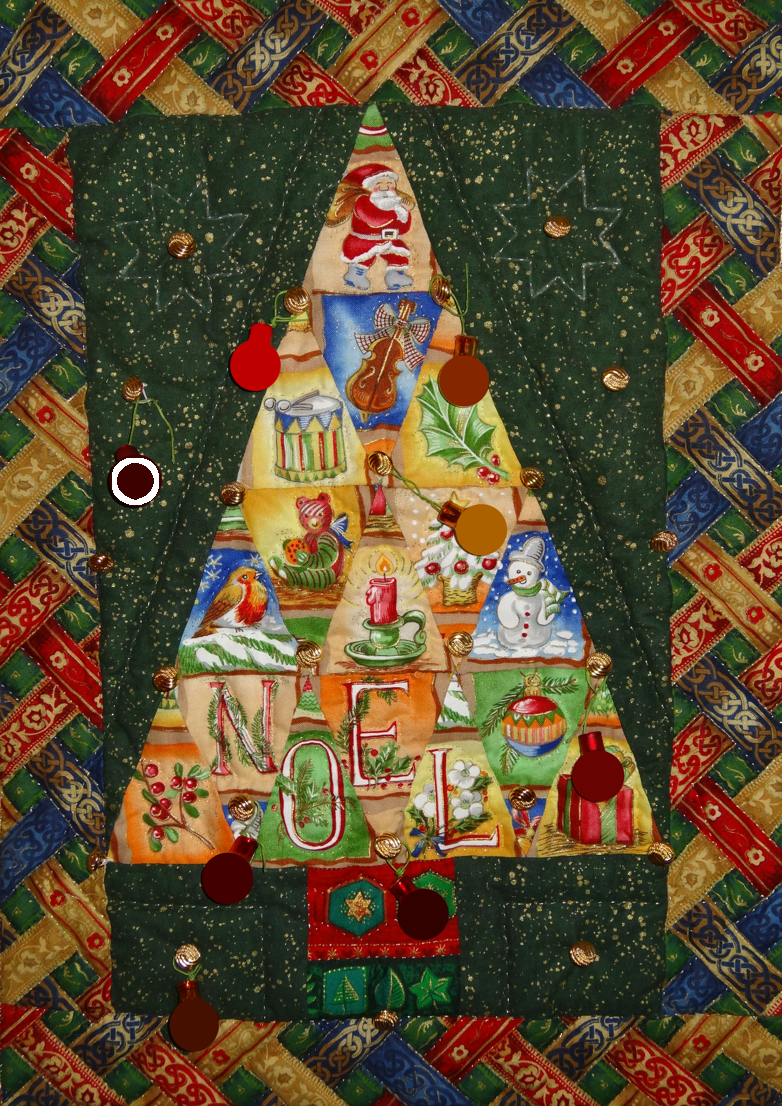}&
	\includegraphics[width=0.29\columnwidth]{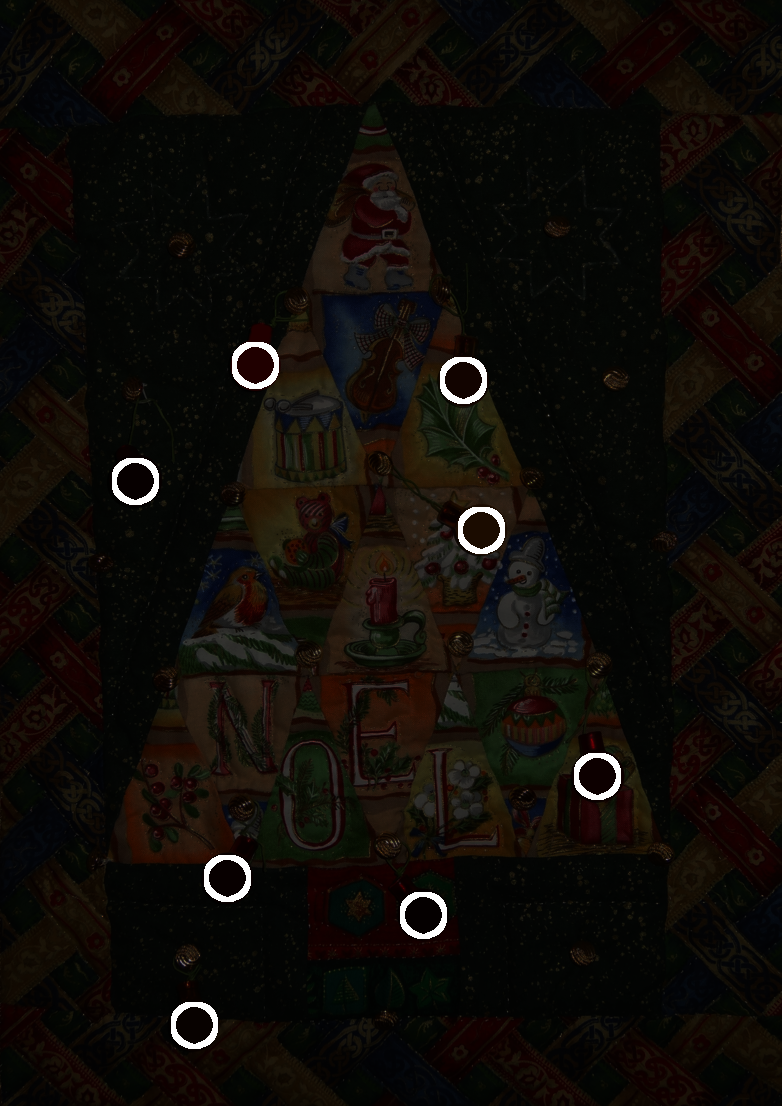}&
	\includegraphics[width=0.29\columnwidth]{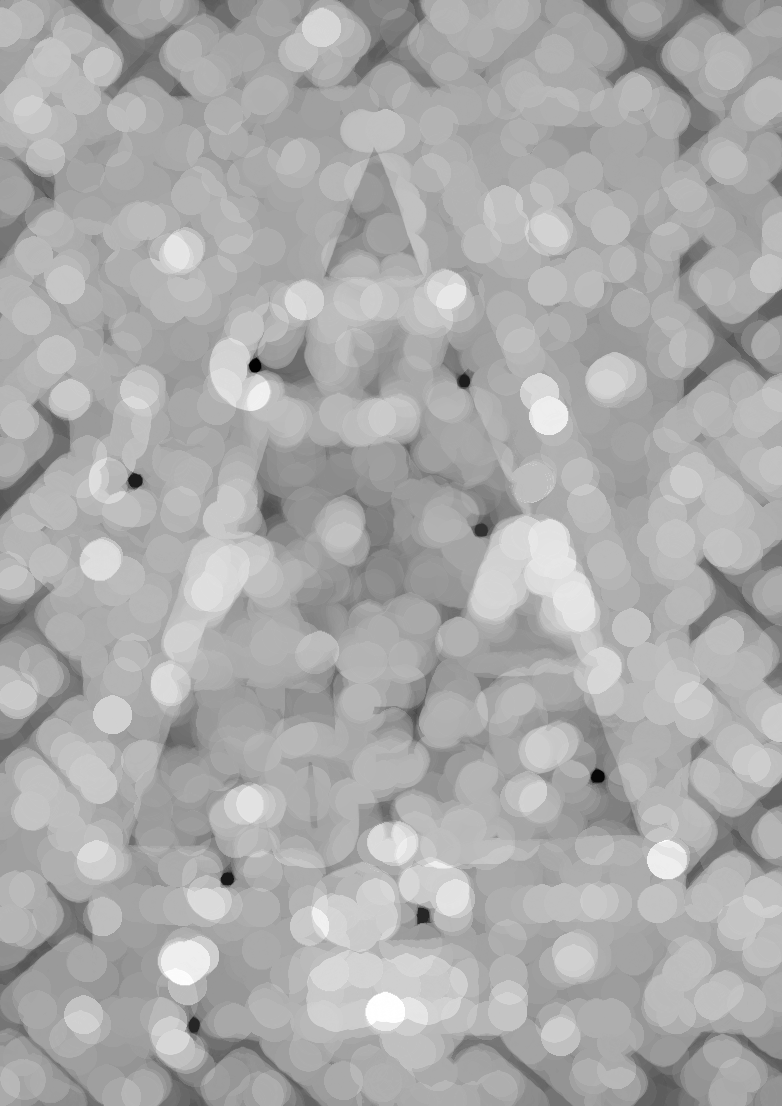}\\
(a) Initial image $\mathbf{f}$&
(b) Dark image $\tilde{\mathbf{f}}$&
(c) Map $As_{\mathbf{t}}^{\LIPCtimes}\tilde{\mathbf{f}}$\\
and probe $\mathbf{t}$&
Balls detected&\\
\end{tabular}
\end{tabular}
\caption{Detection of coloured balls on a dark image $\tilde{\mathbf{f}}$ with a probe $\mathbf{t}$ extracted in the bright image $\mathbf{f}$. (a) The border of the probe $\mathbf{t}$ is coloured in white.}
\label{fig:balls}
\end{figure}

In figure \ref{fig:balls}, two images of the same scene, a bright image $\mathbf{f}$ and a dark image $\tilde{\mathbf{f}}$, are acquired with two different exposure times. The probe $\mathbf{t}$ is extracted in the bright image and used to compute the map of Aspl\"und's distance $As_{\mathbf{t}}^{\LIPCtimes}\tilde{\mathbf{f}}$ in the darker image. By finding the minima of the map, all the balls are detected and their contours are added to the image in figure \ref{fig:balls} (b). One can notice that the Aspl\"und's distance is very robust to the lighting variations.

%
%
\section{Conclusion and perspectives}
\label{sec:concl}
A new spatio-colour Aspl\"und's distance based on colour LIPC model has been defined. It is a true colour (i.e. vectorial) metric based on a colour model consistent with the human visual system. It is also consistent with the previous properties given in \cite{Jourlin2014,Noyel2015}. An extension of this metric robust to noise has been presented and illustrated on pattern recognition examples. This double-sided probing distance is efficient for colour pattern matching and performs better than traditional correlation methods. In future work, we will evaluate in details the properties of this colour distance on practical applications (e.g. in medical, remote sensing or industrial images). We will compare it to the marginal colour Aspl\"und's distance and we will study the links between Aspl\"und's probing and mathematical morphology.

%
%

\bibliographystyle{splncs03}
\bibliography{refs}

\end{document}